\documentclass[11pt,a4paper]{article}
\usepackage[hyperref]{acl2020}
\usepackage{times}
\usepackage{latexsym}

\usepackage{amsmath}
\usepackage{amssymb}
\usepackage{amsfonts}
\usepackage{enumitem}
\usepackage{graphicx}
\graphicspath{{Figure/}}
\usepackage{microtype}

\aclfinalcopy 
\def\aclpaperid{480} 

\title{PLATO: Pre-trained Dialogue Generation Model with \\
Discrete Latent Variable}

\author{Siqi Bao\thanks{~~First two authors contributed equally to this work.}, Huang He\footnotemark[1], Fan Wang, Hua Wu \and Haifeng Wang \\
  Baidu Inc., China \\
  \texttt{\{baosiqi, hehuang, wangfan04, wu\_hua, wanghaifeng\}@baidu.com} \\
  }
\date{}

\begin{document}
\maketitle
\begin{abstract}
    Pre-training models have been proved effective for a wide range of natural language processing tasks. Inspired by this, we propose a novel dialogue generation pre-training framework to support various kinds of conversations, including chit-chat, knowledge grounded dialogues, and conversational question answering. In this framework, we adopt flexible attention mechanisms to fully leverage the bi-directional context and the uni-directional characteristic of language generation. We also introduce discrete latent variables to tackle the inherent one-to-many mapping problem in response generation. Two reciprocal tasks of response generation and latent act recognition are designed and carried out simultaneously within a shared network. Comprehensive experiments on three publicly available datasets verify the effectiveness and superiority of the proposed framework.
\end{abstract}

\section{Introduction}
	Dialogue generation is a challenging task due to the limited corpus of human conversations, complex background knowledge, and diverse relationships between utterances. Recently, pre-trained large-scale language models, such as BERT \cite{devlin2019bert} and XLNet \cite{yang2019xlnet}, have achieved prominent success in natural language processing. Such models are usually constructed based on a massive scale of general text corpora, like English Wikipedia or BooksCorpus \cite{zhu2015aligning}, where distributed representations can be learned automatically from the raw text. With these representations being fine-tuned, breakthroughs have been continuously reported for various downstream tasks, especially those on natural language understanding, such as question answering, natural language inference, and so on.
	
	This pre-training and fine-tuning paradigm also sheds light on the tasks of natural language generation, like dialogue generation. However, the previous study demonstrates that there are some deficiencies in performance while directly fine-tuning BERT on small conversation datasets \cite{rashkin2019towards, wolf2019transfertransfo}. Possible reasons are three-fold: 1) the underlying linguistic patterns in human conversations can be highly different from those in general text, which suggests a potentially large gap in knowledge or data distribution; 2) the training mode of uni-directional dialogue generation is also distinct from that of bi-directional natural language understating as applied in BERT; 3) unlike most of the general NLP tasks, there exists a one-to-many relationship in dialogue generation, where the dialogue context may correspond to multiple appropriate replies.
	
	In this paper, we propose a new method to tackle the above challenges, aiming to obtain a high-quality pre-training model for dialogue generation. First of all, to reduce the gap between data distributions, large-scale Reddit and Twitter conversations are utilized to further pre-train the generation model (upon the basis of language models pre-trained with general text). Secondly, to mitigate the difference in training mode, a flexible paradigm integrating uni- and bi-directional processing is employed in this work, which is inspired by the latest unified language modeling \cite{dong2019unified}. Thirdly, a discrete latent variable is introduced to model the one-to-many relationship among utterances in conversations. Each value of the latent variable corresponds to the particular conversational intent of one response, which is referred as \textit{latent speech act}. 
	
	Distinct with those controllable dialogue generation based on explicit labels (including emotion, keywords, domain codes, and so on) \cite{huang2018automatic, keskar2019ctrl}, our latent variable gets exempted from the restriction of human annotations and can be learned automatically from the corpus in an unsupervised way. In the pre-training of dialogue generation, response generation and latent act recognition are carried out simultaneously within a shared network. Based on the context and latent variable, the generation task tries to maximize the likelihood of the target response. Meanwhile, the recognition task aims to estimate the latent variable w.r.t. the given context and target response. Apparently, the accurate recognition of the latent variable is a crucial factor in boosting the quality of response generation.
	
	We conducted experiments on three different kinds of conversation tasks: chit-chat, knowledge grounded conversation, and conversational question answering. Experimental results verify the effectiveness and superiority of our pre-trained model as compared with the other state-of-the-art methods. Our pre-trained models and source code have been released at GitHub, hoping to facilitate further research progress in dialogue generation.\footnotemark[1]
	\footnotetext[1]{\url{https://github.com/PaddlePaddle/Research/tree/master/NLP/Dialogue-PLATO}}
	
	\section{Dialogue Generation Pre-training}
	Given a piece of context, there exist multiple appropriate responses, leading to diverse conversation flows. It is widely recognized that the capability of modeling one-to-many relationship is crucial for the dialogue generation system \cite{zhao2017learning, chen2019generating}. To this end, we propose to encode discrete latent variables into transformer blocks for one-to-many relationship modeling, where two reciprocal tasks of response generation and latent act recognition are collaboratively carried out.
	
	\subsection{Model Architecture}
	In our model, there are three elements: dialogue context $c$, response $r$ and latent variable $z$. The dialogue context $c$ consists of several history utterances. (For knowledge grounded conversation, it is conventional to concatenate background knowledge into the context as well \cite{wolf2019transfertransfo}.) The response $r$ is one piece of appropriate reply towards the given context. The latent variable $z$ is one $K$-way categorical variable $z\in [1, K]$, with each value corresponding to a particular latent speech act in the response.
	
\begin{figure}
		\centering
		\includegraphics[width=0.43\textwidth]{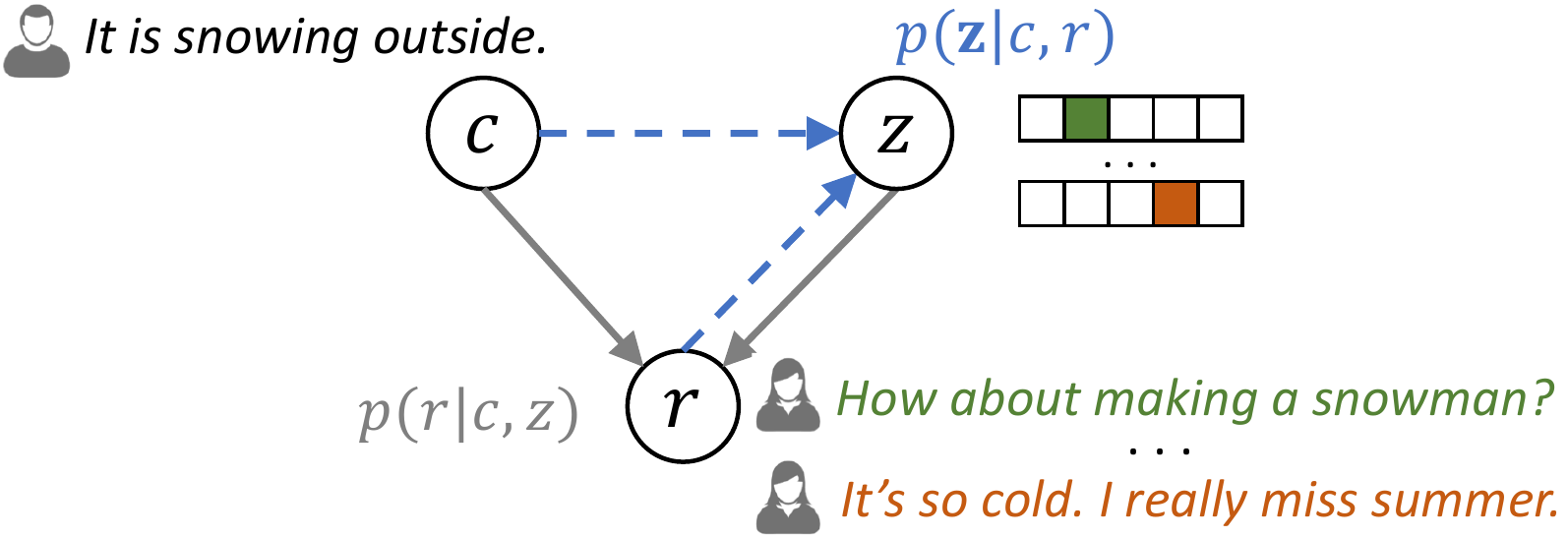}
		\caption{Graphical illustration of response generation (gray lines) and latent act recognition (dashed blue lines).}
		\label{fig:graphical}
	\end{figure}

	The probabilistic relationships among these elements are elaborated with the graphical model in Figure \ref{fig:graphical}. Given a context $c$, there are multiple latent speech acts which can be taken as response intents (represented by the latent variable $z$). Conditioned on the context and one selected latent speech act, the response is generated as $p(r|c,z)$ (gray lines). Given a pair of context and response, the underlying latent speech act can be estimated as $p(z|c,r)$ (dashed blue lines). As such, our pre-training of dialogue generation contains the following two tasks -- \textbf{response generation} and \textbf{latent act recognition}. 

\begin{figure*}[ht]
		\centering
		\includegraphics[width=\textwidth]{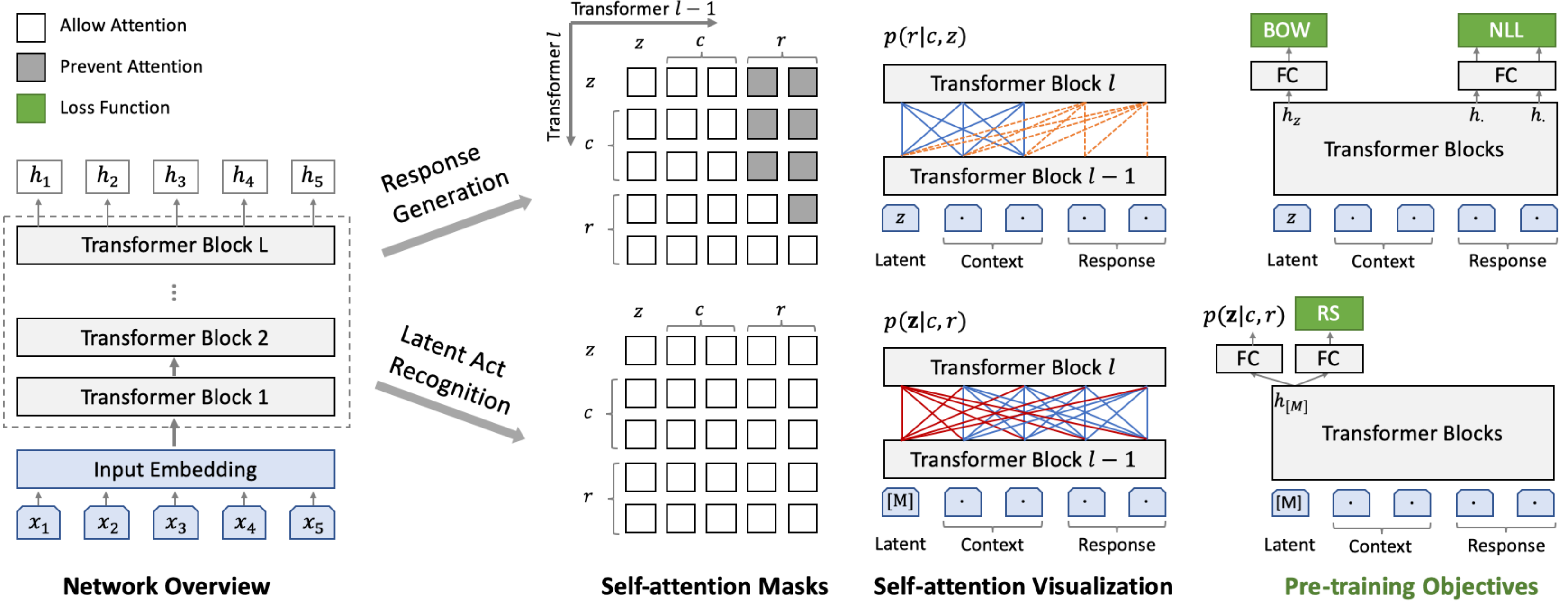}
		\caption{Architecture of dialogue generation with discrete latent variable. In self-attention visualization, \textcolor{red}{red} and \textcolor{blue}{blue} lines denote bi-directional attention, and dashed \textcolor{orange}{orange} lines denote uni-directional attention.}
		\label{fig:Architecture}
	\end{figure*}

	We propose a unified infrastructure for the joint learning of both tasks, shown as Figure \ref{fig:Architecture}. The backbone of our infrastructure is inspired by the transformer blocks in \cite{dong2019unified}, which supports both bi-directional encoding and uni-directional decoding flexibly via specific self-attention masks. Both response generation and latent act recognition are carried out under the unified network with shared parameters. Their detailed implementations are described as follows.
	
	Given the context $c$ and a specific speech act $z$, the response generation can be estimated as
	\begin{equation}
	p(r|c,z)=\Pi_{t=1}^T~p(r_t|c,z,r_{<t})~,
	\end{equation}
	where $T$ is the length of the target response $r$ and $r_{<t}$ denotes previously generated words. Since the response generation is a uni-directional decoding process, each token in the response only attends to those before it, shown as dashed orange lines in Figure \ref{fig:Architecture}. 

	\begin{figure*}[ht]
		\centering
		\includegraphics[width=\textwidth]{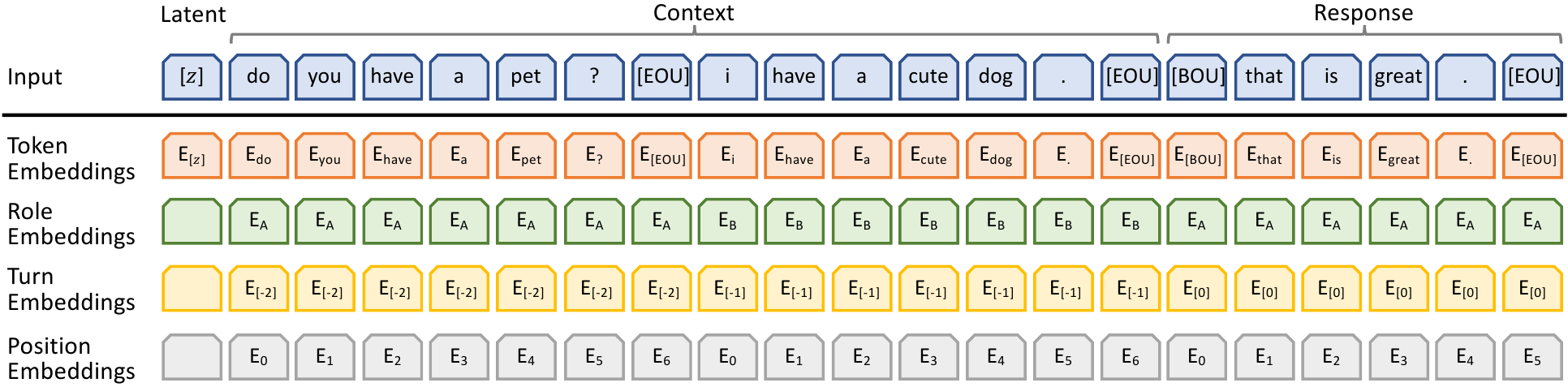}
		\caption{Input representation. The input embedding is the sum of token, role, turn and position embeddings.}
		\label{fig:input}
	\end{figure*} 
	The latent act recognition task is included to identify the corresponding value of $z$ for the given context and the target response in the training data. The latent act recognition shares network parameters with response generation, but has a separate self-attention mask for bi-directional encoding. As shown in Figure \ref{fig:Architecture}, with a special mask symbol [M] as input, it keeps collecting information from the context and target response (red lines). In this way, the corresponding speech act for the target response can be recognized as $z\sim p(\mathbf{z}|c,r)$, where $p(\mathbf{z}|c,r)$ is the estimated posterior distribution over discrete latent values. 
	
	\subsection{Input Representation}
	For multi-turn conversation modeling, elaborate designs have been made on the input representation in this work. For each token, its input embedding is the sum of corresponding token, role, turn and position embeddings. One visual example is shown in Figure \ref{fig:input} and details are described in the following.
	\begin{itemize}[leftmargin=*,noitemsep,topsep=0pt]
		\item The input is the concatenation of latent variable, dialogue context and response. Following the pre-processing of BERT \cite{devlin2019bert}, the input text is tokenized with WordPiece \cite{wu2016google}. A special end-of-utterance [EOU] token is appended to the end of each utterance for separation. Another begin-of-utterance [BOU] token is added at the beginning of the response, whose final hidden state (i.e., output of the last transformer block) is used to predict next token during generation.
		
		\item Given that $z$ is one $K$-way categorical variable, its token embedding $\mathbf{E}_{[z]}$ is mapped from the latent embedding space $\mathbf{E}_\mathbf{z}\in \mathbb{R}^{K\times D}$. For the rest tokens in the vocabulary, they are initialized using BERT's WordPiece embeddings. 
		
		\item Role embeddings are employed to differentiate the characters evolved in the conversation. The role embedding $\mathbf{E}_A$ is added for the response, as well as dialogue utterances generated by the same character in the context. And role embedding $\mathbf{E}_B$ is used for the other character. (For knowledge grounded conversation, $\mathbf{E}_C$ is used as the role embedding of background knowledge.) 
		
		\item In the interactive conversation, there are multi-turn utterances and we employ relative order in the assignment of turn embeddings. The turn embedding for the response is set to $\mathbf{E}_{[0]}$, and the turn embedding of its last utterance is $\mathbf{E}_{[-1]}$, and etc. Our utilization of relative turn embeddings instead of absolute ones enables the model to assign turn embedding $\mathbf{E}_{[0]}$ to the response consistently and makes response generation exempt from the disturbance of its round number within the dialogue.
		
		\item Position embeddings are added according to the token position in each utterance. Note that for the special token of latent variable, its corresponding role, turn and position embeddings are all set to empty. 
	\end{itemize}
	
	\subsection{Pre-training Objectives}
	We employ three loss functions in dialogue generation pre-training: negative log-likelihood (NLL) loss, bag-of-words (BOW) loss and response selection (RS) loss. Brief illustration is shown in the last column of Figure \ref{fig:Architecture} and detailed descriptions will be provided in this section.
	\subsubsection{Response Generation}
	In our model, the response is generated conditioned on the latent variable and the context. The widely adopted NLL loss is embraced in the pre-training:
	\begin{equation}
	\begin{split}
	\mathcal{L}_{NLL}&=-\mathbb{E}_{z\sim p(\mathbf{z}|c,r)} ~\log p(r|c,z)\\
	&=-\mathbb{E}_{z\sim p(\mathbf{z}|c,r)} \sum_{t=1}^T~\log p(r_t|c,z,r_{<t})~,
	\end{split}
	\raisetag{2.6\baselineskip}
	\end{equation}
	where $z$ is the latent speech act of this training pair $(c, r)$, sampled from the probability distribution $p(\mathbf{z}|c,r)$. The posterior distribution over latent values is estimated through the task of latent act recognition: 
	\begin{equation}
	p(\mathbf{z}|c,r) =\text{softmax}(W_1 h_{[M]} + b_1) \in \mathbb{R}^K~,
	\end{equation}
	where $h_{[M]}\in \mathbb{R}^{D}$ is the final hidden state of the special mask, $W_1\in \mathbb{R}^{K\times D}$ and $b_1\in \mathbb{R}^{K}$ denote the weight matrices of one fully-connected layer.
	
	Besides the classical NLL loss, the bag-of-words loss \cite{zhao2017learning} is also employed to facilitate the training process of latent discrete variables:
	\begin{equation}
	\begin{aligned}
	\mathcal{L}_{BOW}&=-\mathbb{E}_{z\sim p(\mathbf{z}|c,r)}  \sum_{t=1}^T ~ \log p(r_t|c,z)\\
	&=-\mathbb{E}_{z\sim p(\mathbf{z}|c,r)}  \sum_{t=1}^T ~ \log \frac{e^{f_{r_t}}}{\sum_{v\in V} e^{f_v}}~,
	\end{aligned}
	\end{equation}
	where $V$ refers to the whole vocabulary. The function $f$ tries to predict the words within the target response in a non-autoregressive way:
	\begin{equation}
	f=\text{softmax} (W_2 h_z+b_2) \in \mathbb{R}^{|V|}~,
	\end{equation}
	where $h_z$ is the final hidden state of the latent variable and $|V|$ is the vocabulary size. $f_{r_t}$ denotes the estimated probability of word $r_t$. As compared with NLL loss, the BOW loss discards the order of words and forces the latent variable to capture the global information of the target response.
	
	\subsubsection{Response Selection}
	Response selection helps distinguish whether the response is relevant with the dialogue context and consistent with the background knowledge. Meanwhile, its score can be regarded as an indicator of coherence during inference, helping to select the most coherent one from multiple candidate responses.
	
	Particularly, the training of response selection is carried out together with the bi-directional encoding  of latent act recognition. The positive training samples come from the dialogue context and corresponding target response $(c,r)$, with label $l_r=1$. And the negative samples are created by randomly selecting responses from the corpus $(c,r^-)$, with label $l_{r^-}=0$. The binary cross-entropy loss of response selection is defined as follows:
	\begin{equation}
	\mathcal{L}_{RS}=-\log p(l_r=1|c,r)-\log p(l_{r^-}=0|c,r^-)
	\end{equation}
	The above probability is estimated through one fully-connected layer, with the final hidden state of the special mask fed as input:
	\begin{equation}
	p(l_r=1|c,r)=\text{sigmoid} (W_3 h_{[M]}+b_3)
	\end{equation}
	
	To sum up, the total objective of our pre-training model is to minimize the integrated loss:
	\begin{equation}\label{eq:obj}
	\mathcal{L}=\mathcal{L}_{NLL}+ \mathcal{L}_{BOW} +	\mathcal{L}_{RS}
	\end{equation}
	
	\subsection{Pre-training Procedure}
	Our pre-training model contains 12 transformer blocks, with network parameters initialized using BERT\textsubscript{BASE}. Large-scale conversation datasets -- Twitter \cite{cho2014learning} and Reddit \cite{zhou2018commonsense, galley2019grounded} are employed for pre-training, which results in 8.3 million training samples in total. For each training sample of context and target response $(c,r)$, it needs to pass through the network twice to accomplish the tasks of latent act recognition and response generation. And the pre-training steps are summarized as follows:
	\begin{enumerate}[label=\arabic*),leftmargin=*,noitemsep,topsep=0pt]
		\item Latent Act Recognition
		\begin{itemize}[leftmargin=*,noitemsep,topsep=0pt]
			\item[--] Given a pair of context and target response, estimate the posterior distribution $p(\mathbf{z}|c,r)$
			\item[--] Randomly select $r^-$ and calculate $\mathcal{L}_{RS}$
		\end{itemize}
		
		\item Response Generation
		\begin{itemize}[leftmargin=*,noitemsep,topsep=0pt]
			\item[--] With the sampled latent value $z\sim p(\mathbf{z}|c,r)$, calculate $\mathcal{L}_{NLL}$ and $\mathcal{L}_{BOW}$
		\end{itemize}
		
		\item Optimization
		\begin{itemize}[leftmargin=*,noitemsep,topsep=0pt]
			\item[--] Sum up to obtain $\mathcal{L}$, and update network parameters with back-propagation
		\end{itemize}
	\end{enumerate}
	
	The hyper-parameters used in pre-training are listed as follows. The maximum sequence length of context and response is set to 256 and 50, respectively. The number of transformer blocks in our model $L$ is 12 and the hidden embedding dimension $D$ is 768. The batch size is set to 64 and $K$ is set to 20 for the discrete latent variable. Adam optimizer \cite{kingma2015adam} is employed for optimization with a learning rate of 5e-5. The pre-training of dialogue generation was carried out on 8 Nvidia Telsa V100 32G GPU cards for 3.5M steps, taking about two weeks to reach convergence. 
	
	\subsection{Fine-tuning and Inference}
	Our pre-trained model is flexible enough to support various kinds of dialogues, including chit-chat, knowledge grounded conversation, conversational question answering, etc. The fine-tuning on small conversation datasets can be carried out by following the training objectives defined in Equation \eqref{eq:obj}. As the fine-tuning process reaches convergence, the response towards the given context can be obtained through the following inference procedure:
	\begin{enumerate}[label=\arabic*),leftmargin=*,noitemsep,topsep=0pt]
		\item Candidate Response Generation
		\begin{itemize}[leftmargin=*,noitemsep,topsep=0pt]
			\item[--] Conditioned on each latent value $z\in [1, K]$, generate corresponding candidate response $r$.
		\end{itemize}
		
		\item Response Selection
		\begin{itemize}[leftmargin=*,noitemsep,topsep=0pt]
			\item[--] Calculate the probability for each response $p(l_r=1|c,r)$ and select the one with highest coherence value as the final response.
		\end{itemize}
	\end{enumerate}

	It is worth noting that the above fine-tuning and inference procedures are set up for the dialogue generation without any specific objectives. If there exists a specific objective within the conversation, such as letting both participants know more about each other \cite{bao2019know}, the fine-tuning can proceed to maximize the pre-defined rewards with reinforcement learning (RL). Under such circumstances, our latent discrete variable can be naturally treated as action within RL, and thus the response selection can be straightforwardly solved by selecting the action that results in the maximum reward. 
	 
	\section{Experiments}
	\subsection{Settings}
	\subsubsection{Datasets}
	To evaluate the performance of our proposed method, comprehensive experiments have been carried out on three publicly available datasets. 
	\begin{itemize}[leftmargin=*,noitemsep,topsep=0pt]
		\item Persona-Chat \cite{zhang2018personalizing} is a knowledge grounded conversation dataset. It provides both manually annotated conversations and corresponding persona profiles (background knowledge), where two participants chat naturally and try to get to know each other.
		
		\item Daily Dialog \cite{li2017dailydialog} is a chit-chat dataset, which contains high-quality human conversations about daily life. 
		
		\item DSTC7-AVSD \cite{alamri2019audio}, short for Audio Visual Scene-aware Dialog of the DSTC7 challenge, is a conversational question answering dataset. In DSTC7-AVSD, the system need to generate an answer given dialogue context and background knowledge. There are two available options of knowledge utilization: 1) using single-modal information of text only, including video's caption and summary; 2) relying on multi-modal information, including text, audio and visual features. The single-modal option is adopted by our method in the experiments.
	\end{itemize}
	The descriptions and statistics of these datasets are summarized in Table \ref{tab:dataset}. 

	\subsubsection{Compared Methods}
	The following models have been compared in the experiments.
	
	\noindent \textbf{Baseline.} Sequence to sequence with attention (Seq2Seq) \cite{vinyals2015neural} is employed as the baseline for the experiments on Persona-Chat and Daily Dialog. DSTC7-AVSD has provided a baseline system, which is built upon hierarchical recurrent encoders with multi-modal features.
		
	\noindent\textbf{State of the art.} Persona-Chat was also utilized in the ConvAI2 challenge \cite{dinan2019second}, where the team of Lost in Conversation (LIC) \cite{golovanov2019large} obtains the best performance. LIC is also one transformer based generation method and fine-tuned upon the pre-trained model of GPT \cite{radford2018improving}. For the dataset of Daily Dialog, its best results are reported by the recently developed method -- iVAE\textsubscript{MI} \cite{fang2019implicit}, which generates diverse responses with sample-based latent representation. In DSTC7-AVSD, the team of CMU \cite{sanabria2019cmu} obtains the best performance across all the evaluation metrics. 
		
	\noindent \textbf{Our method.} To better analyze the effects of our latent discrete variable, we also compare to the version without latent variable (Our w/o Latent).\footnotemark[2]
	\footnotetext[2]{It shares the same training settings as our method with latent variables: network parameters are first initialized with BERT\textsubscript{BASE}, and the pre-training is further carried out on Reddit and Twitter. The only difference lies in the incorporation of latent variable.}
	
	\begin{table*}
		\centering
		\includegraphics[width=\textwidth]{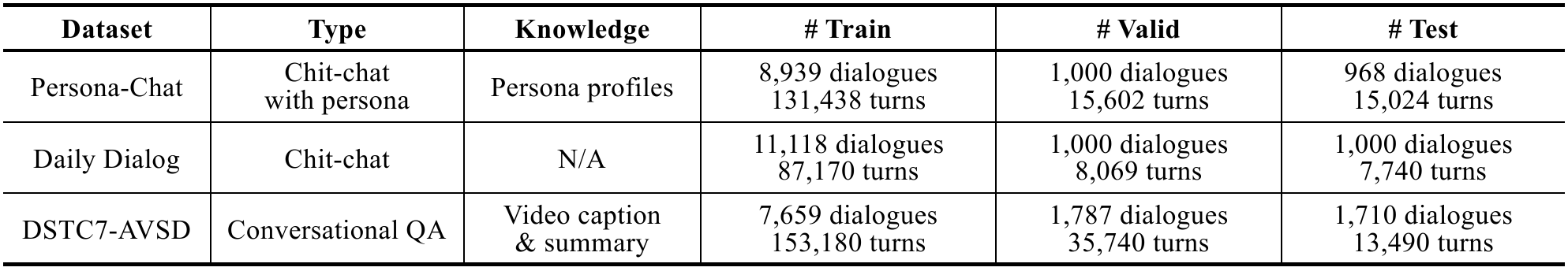}
		\caption{Summary of datasets used in the experiments.}
		\label{tab:dataset}
	\end{table*} 
	\begin{table*}
		\centering
		\includegraphics[width=\textwidth]{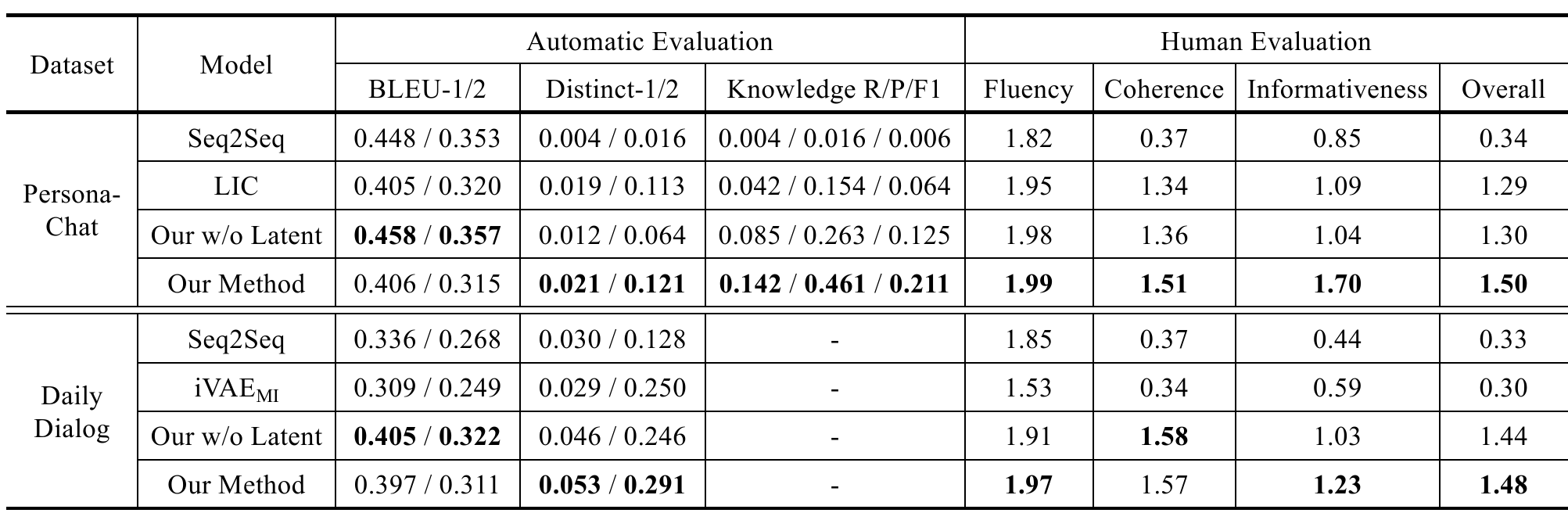}
		\caption{Experimental results on Persona-Chat and Daily Dialog with automatic and human evaluations, with highest value written in bold.}
		\label{tab:conv}
	\end{table*} 
	\begin{table*}
		\centering
		\includegraphics[width=\textwidth]{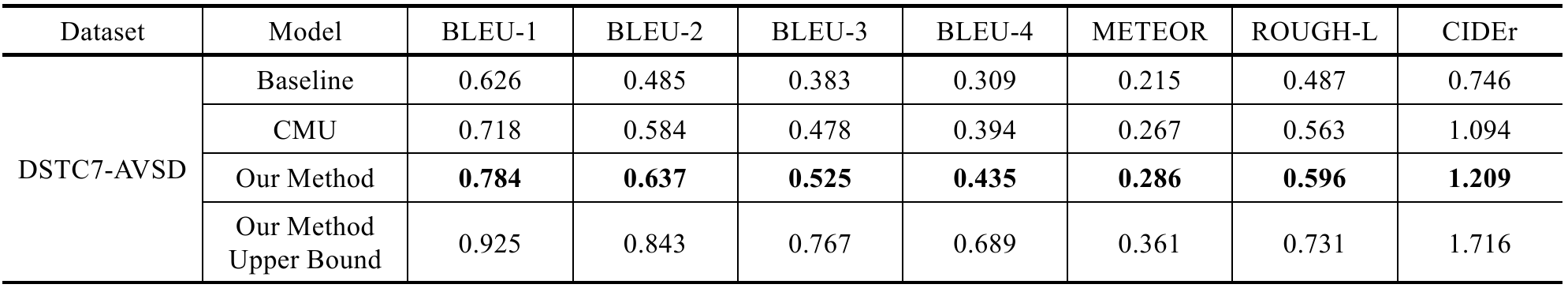}
		\caption{Experimental results on DSTC7-AVSD with automatic evaluation, with highest value written in bold.}
		\label{tab:qa}
	\end{table*} 
	\begin{table*}
		\centering
		\includegraphics[width=\textwidth]{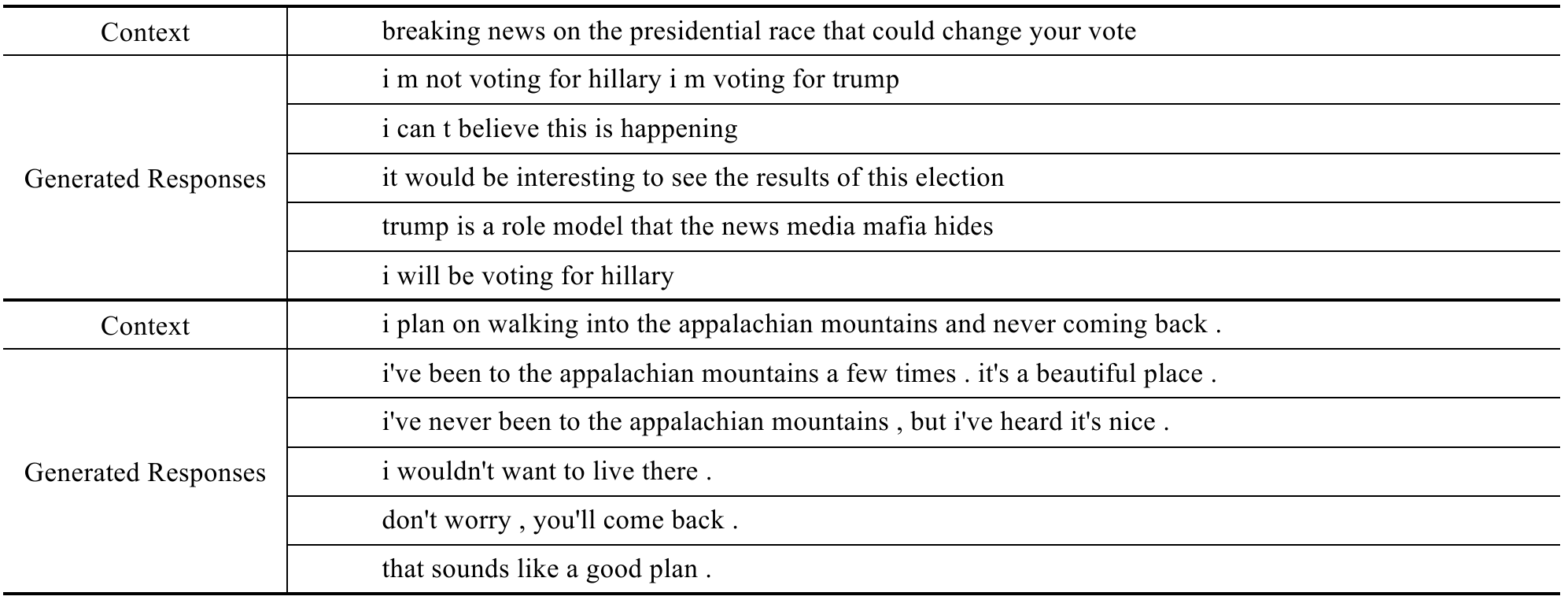}
		\caption{Examples of response generation with our pre-trained model.}
		\label{tab:case_1}
	\end{table*}
	\begin{table*}
		\centering
		\includegraphics[width=\textwidth]{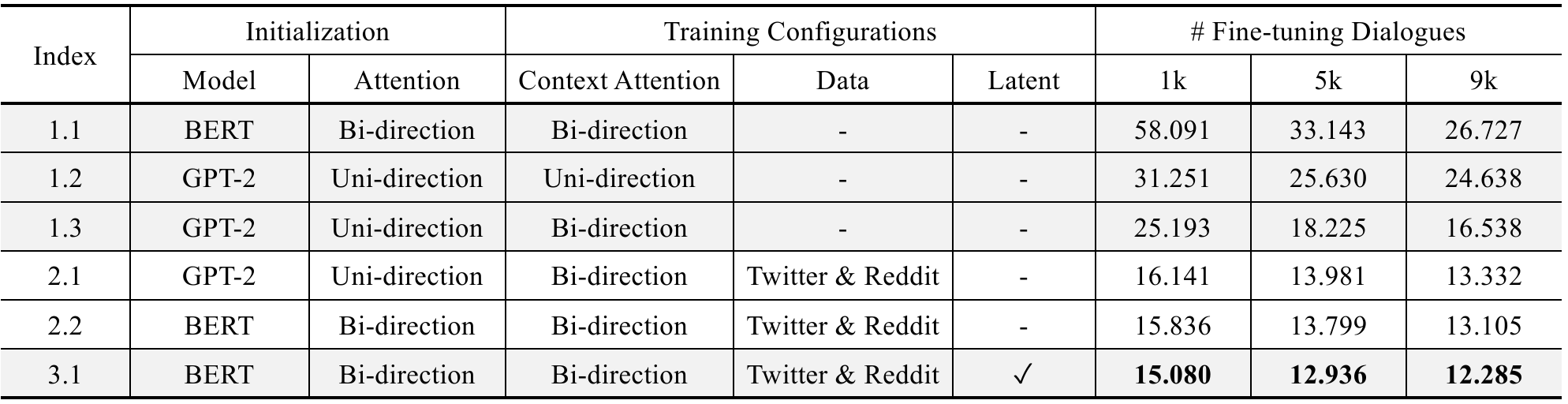}
		\caption{Perplexity of different pre-trained models on Persona-Chat, with best value written in bold.}
		\label{tab:trans}
	\end{table*}
	
	\subsubsection{Evaluation Metrics}
	Both automatic and human evaluations are employed to assess the performance of compared methods. In automatic evaluation, the following metrics are included:
	\begin{itemize}[leftmargin=*,noitemsep,topsep=0pt]
		\item BLEU \cite{chen2014systematic} measures the n-gram overlap between generated response and the target response.
		
		\item Distinct-1/2 \cite{li2016diversity} measures the generation diversity, which is defined as the number of distinct uni- or bi-grams divided by the total amount of generated words. 
		
		\item Knowledge Recall/Precision/F1 \cite{dinan2019wizard} measures the degree of informativeness w.r.t. background knowledge. 
		
		\item In DSTC7-AVSD, the MSCOCO platform \cite{chen2015microsoft} is employed for evaluation. It compares the generated response with six ground truth responses, using metrics of BLEU, METEOR, ROUGH-L and CIDEr. 
		\end{itemize}

	In human evaluation, we randomly select 100 dialogue contexts and generate responses with compared methods. Three crowd-sourcing workers are asked to score the response quality on a scale of [0, 1, 2] from four aspects -- fluency, coherence, informativeness and overall. The higher score, the better. Details about the criteria are given as follows.
	\begin{itemize}[leftmargin=*,noitemsep,topsep=0pt]
		\item Fluency measures whether the generated sentence is smooth and grammatically correct.
		
		\item Coherence evaluates whether the generated response is relevant with the dialogue context and consistent with the expressed information or background knowledge.
		
		\item Informativeness assesses whether the response is informative or not.
		
		\item Overall represents the general evaluation, where 0 indicates a bad response, 1 refers to a normal response and 2 stands for a good response.
	\end{itemize}
	After collecting the assessments from annotators, the response's final score is determined via majority voting. The average Fleiss's kappa \cite{fleiss1973equivalence} on Persona-Chat and Daily Dialog is 0.515 and 0.480 respectively, indicating annotators have reached moderate agreement.
 
	\subsection{Experimental Results}
	The experimental results on Persona-Chat and Daily Dialog with automatic and human evaluations are summarized in Table \ref{tab:conv}. 
	As suggested in the empirical study \cite{liu2016not}, the correlation between automatic metrics and human judgments is weak in open-domain dialogue generation. In the automatic evaluation, experimental results demonstrate that no method can consistently outperform the others.
	
	During human evaluations, our method achieves better performance consistently across all the metrics on Persona-Chat and Daily Dialog. The scores of fluency almost approach the upper bound, revealing that our generated responses are very fluent. The informativeness assessments indicate that the information in our generated responses is significantly richer, as compared with the baseline methods. Our responses are coherent with the context and favored most by crowd-sourcing workers. The ablation study with our method and our w/o latent also suggests that through the incorporation of discrete latent variables, remarkable improvements can be achieved for dialogue generation. In addition, it can be observed that the generation quality of transformed-based approaches (LIC and our method) is significantly better than that of RNN-based methods (Seq2Seq and iVAE\textsubscript{MI}).\footnotemark[3]
	\footnotetext[3]{It is a normal phenomenon that the performance of our w/o latent is close to that of LIC. Both of them initialize network parameters with pre-trained language models and continue training with large-scale conversation data as Reddit.}
	
	The experimental results on DSTC7-AVSD with automatic evaluation are provided in Table \ref{tab:qa}. 
	In the experiments, our response selection is strengthened with an extra ranking step, which learns to rank the candidates according to automatic scores and selects the top one as the final answer. The results in Table \ref{tab:qa} demonstrate that our method has brought a new breakthrough for DSTC7-AVSD. Additionally, the upper bound of our method is also reported, under the ideal scenario that the optimal candidate answer can be selected.\footnotemark[4] The incredible results validate the great potential of our approach.
	\footnotetext[4]{Given a dialogue context and background knowledge, our model is able to generate $K$ diverse responses. Each of them will be evaluated using MSCOCO and the one obtaining the best score will be treated as the optimal candidate answer.}
	
	\subsection{Discussions}
	\subsubsection{Case Analysis}
	To further dissect the quality of our pre-trained model, several examples of generated responses are provided in Table \ref{tab:case_1}. For each piece of context, our model can produce multiple responses by assigning distinct values to the latent variable and five candidate responses are selected for display in the table. It shows that our pre-trained model is able to generate diverse and appropriate responses. More examples on the conversational datasets are provided in the Appendix.
	
	\subsubsection{Comparison of Pre-trained Models}
	To further analyze the effectiveness of our pre-trained model, more ablation studies have been conducted on Persona-Chat. Distinct pre-trained models are included for comparison. To be fair, their transformer layers are all set to 12. There are three different sizes of training dialogues: 1k, 5k and 9k (all training data). The training configurations and experimental results measured with perplexity are summarized in Table \ref{tab:trans}. There are three groups of pre-trained models: group 1 applies direct fine-tuning of BERT or GPT-2 \cite{radford2019language} on Persona-Chat; group 2 employs Twitter and Reddit for further training upon the basis of pre-trained language models; group 3 carries out the training process with latent variable.\footnotemark[5] (Model 2.2 is our w/o latent one and model 3.1 is our method.) 
	\footnotetext[5]{Overall, group 1 involves two-stage training: pre-training of language model with general text and fine-tuning on small conversation datasets. Whereas, group 2 and group 3 involve three-stage training: pre-training of language model with general text, further pre-training of dialogue generation with Twitter and Reddit, and fine-tuning on small conversation datasets.}
	
	These results demonstrate that our method outperforms the other pre-trained models consistently with lower perplexity across different training sets. Several interesting conclusions can be also drawn from these results. Firstly, the comparison between model 1.2 and model 1.3 encourages the adoption of flexible attention mechanism to fully leverage the bi-directional context information.\footnotemark[6] Secondly, the superiority of group 2 over group 1 mainly comes from the employment of Twitter and Reddit, which are closer to human conversations than general text. Thirdly, the comparison between model 2.2 and model 3.1 reflects that the incorporation of discrete latent variable is able to boost the quality of response generation, whose effects have also been verified in Table \ref{tab:conv}.
	\footnotetext[6]{The results of model 1.1 demonstrate that there are some deficiencies in performance to apply direct fine-tuning of BERT on small conversation datasets, as discussed in the introduction.}

	\section{Related Work}
	Related work involves pre-trained language models and one-to-many modeling in dialogue generation. 
	
	\noindent\textbf{Pre-trained Language Models.} Pre-trained language models, which are trained on massive general text, have brought many breakthroughs on various NLP tasks. These models can be roughly divided into two categories according to their attention mechanisms. GPT \cite{radford2018improving} and GPT-2 \cite{radford2019language} are representative uni-directional language models, where one token is only allowed to attend its previous tokens and the objective is to maximize left-to-right generation likelihood. BERT \cite{devlin2019bert} and XLNet \cite{yang2019xlnet} are bi-directional language models, where bi-directional context attention is enabled for token prediction. The latest unified language model UniLM \cite{dong2019unified} is able to support both uni- and bi-directional attention with flexible self-attention mask designs. Recently, some attempts \cite{golovanov2019large, wolf2019transfertransfo, zhang2019dialogpt} have been made to adapt generative language models GPT or GPT-2 for dialogue generation. Whereas the special issues of conversations, such as impacts from background knowledge and problems of one-to-many relationship, are not fully considered and tackled in these adaptations. 
	
	\noindent\textbf{One-to-many Modeling.} Given one piece of context, there exists multiple appropriate responses, which is know as the one-to-many mapping problem. To model this one-to-many relationship, CVAE \cite{zhao2017learning} employs Gaussian distribution to capture the discourse-level variations of responses. To alleviate the issue of posterior collapse in VAE, some extension approaches are further developed, including conditional Wasserstein auto-encoder of DialogWAE \cite{gu2019dialogwae} and implicit feature learning of iVAE\textsubscript{MI} \cite{fang2019implicit}. SpaceFusion\cite{gao2019jointly} aims to jointly optimize diversity and relevance in the latent space, which are roughly matched by the distance and direction from the predicted response vector. Besides the continuous representation in VAE, discrete categorical variables are also utilized for interpretable generation \cite{zhao2018unsupervised}. Additionally, multiple mapping modules as latent mechanisms are introduced for diverse generation \cite{chen2019generating}, where accurate optimization is carried out via posterior mapping selection. However, due to the small scale of annotated conversation data and limited capacity of generation network, it remains challenging for these methods to balance the diversity and fluency during response generation. 
	
	\section{Conclusion}
	A novel pre-training model for dialogue generation is introduced in this paper, incorporated with latent discrete variables for one-to-many relationship modeling. To pre-train our model, two reciprocal tasks of response generation and latent recognition are carried out simultaneously on large-scale conversation datasets. Our pre-trained model is flexible enough to handle various down-stream tasks of dialogue generation. Extensive and intensive experiments have been carried out on three different kinds of publicly available datasets. And the results demonstrate that our model obtains significant improvements over the other state-of-the-art methods.
	
	Our work can be potentially improved with more fine-grained latent variables. In the future, we will also explore to boost the latent selection policy with reinforcement learning and extend our pre-training to support dialogue generation in other languages.

\section*{Acknowledgments}
We would like to thank the ACL reviewers for their constructive suggestions and Chaotao Chen, Junkun Chen, Tong Wu and Wenxia Zheng for their generous help. This work was supported by the Natural Key Research and Development Project of China (No. 2018AAA0101900).

\bibliography{bibtex}
\bibliographystyle{acl_natbib}

\appendix
\section{Additional Case Analysis}
	In Table \ref{tab:case_2}, it provides the cases of our method and compared approaches on Persona-Chat, where two participants chat with each other according to their personas. As shown in the example, participant P2 needs to produce a response towards the given dialogue context, conditioned on his/her persona profile. The baseline Seq2Seq tends to generate common replies with low informativeness and poor coherence. LIC and our w/o latent are able to produce some coherent responses, whereas deficient in informativeness. In comparison, the response by our method is not only coherent with the context, but also expressive of the background personas. 
	Besides, we also observe the phenomenon of diverse knowledge usage in our response generation, which suggests that the latent variable helps control the knowledge selection and utilization in an implicit way.
	
	Table \ref{tab:case_3} provides the generated responses on Daily Dialog, where two participants chat about daily life. This example shows that Seq2Seq is able to generate fluent utterances, while lacking coherence with the context. As for iVAE\textsubscript{MI}, it suffers from the difficulty to balance diversity and fluency. By contrast, our method is able to generate more coherent and high-quality responses.
	
	Table \ref{tab:case_4} provides the generated responses on DSTC7-AVSD, where two participants discuss the objects and events in a video. Participant P1 is responsible to raise questions, who only has access to the first, middle and last frames of the video. Participant P2 has watched the whole video and needs to answer the partner's questions. The generation system is developed to mimic P2 and answer the questions based on the background knowledge. The baseline approach relies on the multi-modal information, including text, audio and visual features, to produce the answer. Due to the limited network capacity, it lacks fidelity to the background knowledge and makes the generated response less accurate. (As the team of CMU has not released their codes or models, their samples are omitted here.) Our method utilizes the video caption and summary as background knowledge. It can be observed that our method generates a more appropriate answer as compared with the baseline approach. 
	
	\begin{table*}
		\centering
		\includegraphics[width=\textwidth]{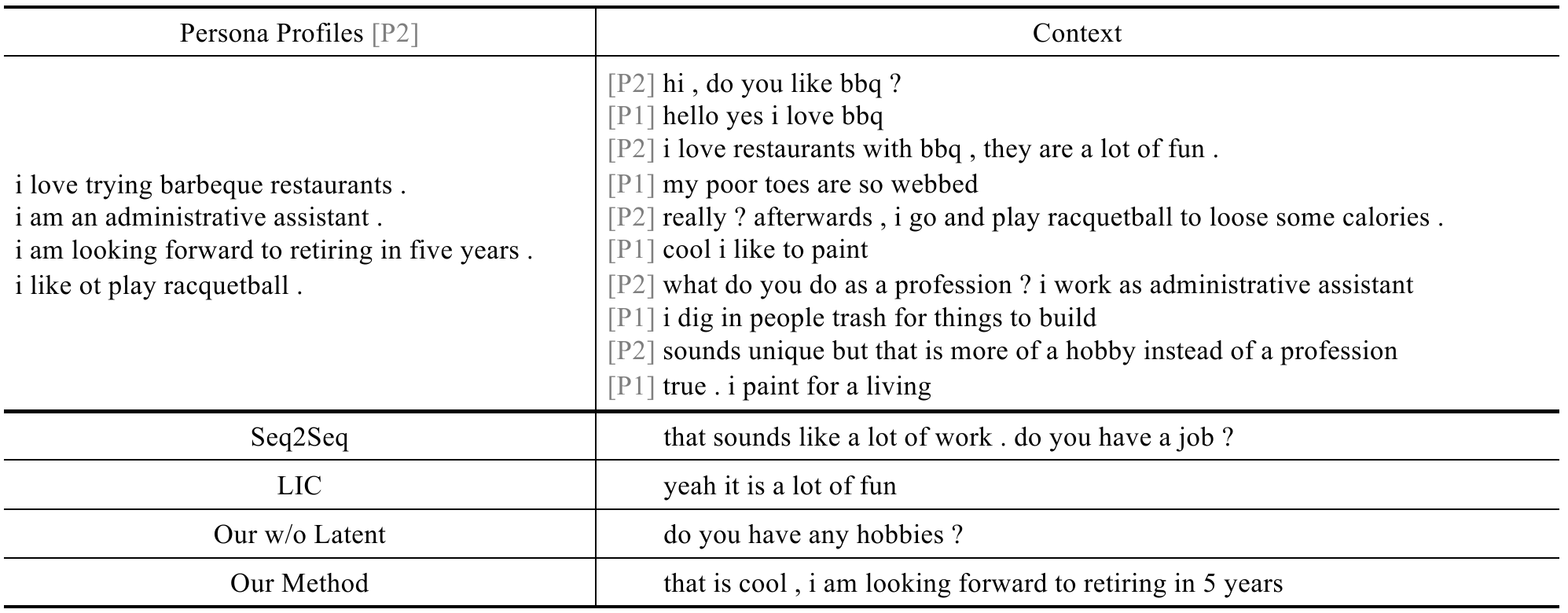}
		\caption{Case analysis of response generation on Persona-Chat.}
		\label{tab:case_2}
	\end{table*} 
	\begin{table*}
		\centering
		\includegraphics[width=\textwidth]{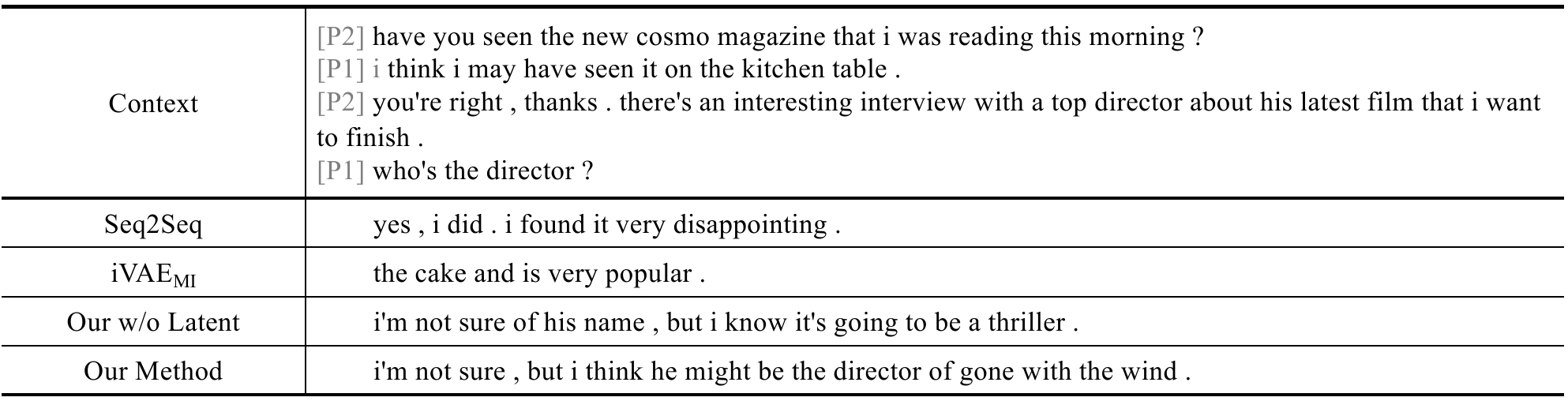}
		\caption{Case analysis of response generation on Daily Dialog.}
		\label{tab:case_3}
	\end{table*} 
	\begin{table*}
		\centering
		\includegraphics[width=\textwidth]{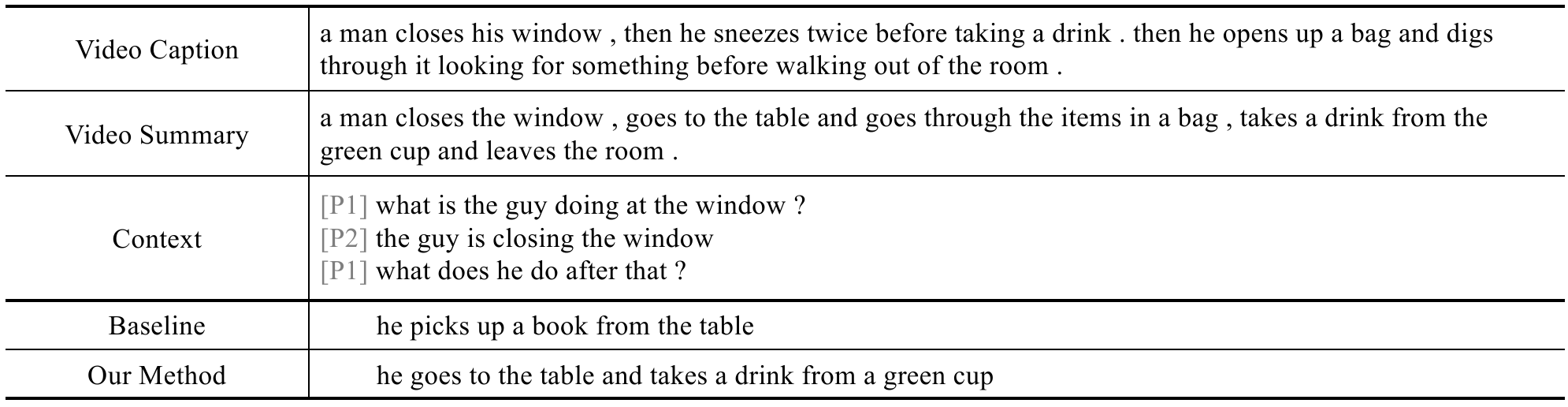}
		\caption{Case analysis of response generation on DSTC7-AVSD.}
		\label{tab:case_4}
	\end{table*} 

\end{document}